\pdfoutput=1 

\documentclass[11pt,a4paper]{article}
\usepackage[hyperref]{emnlp2020}
\usepackage{times}
\usepackage{latexsym}

\usepackage{microtype}
\usepackage{mathtools}
\aclfinalcopy 
\def\aclpaperid{1612} 

\usepackage{kotex}
\usepackage{adjustbox}
\usepackage{enumitem}
\usepackage{xcolor}
\usepackage{multirow}
\usepackage{graphicx}
\usepackage{natbib}

\usepackage{amsmath}
\newcommand{\abs}[1]{\left\lvert #1 \right\rvert}
\usepackage{amssymb}
\DeclarePairedDelimiterX{\Set}[2]\{\}{%
  \, #1 \;\delimsize\vert\; #2 \,
}
\usepackage{lipsum}
\usepackage{algorithm}
\usepackage{algpseudocode}

\newcommand{\xti}{\tilde{x}_i}
\newcommand{\xtj}{\tilde{x}_j}
\usepackage{lipsum}
\usepackage{algorithm,algpseudocode}
\usepackage{multicol} 
\usepackage{tabularx}
\usepackage{booktabs}
\usepackage{rotating}
\usepackage{multirow}

\newcommand{\red}[1]{\textcolor{black}{#1}}

\aclfinalcopy

\title{Interpretation of {NLP} models through input marginalization}

\author{Siwon Kim \quad Jihun Yi \quad Eunji Kim \quad Sungroh Yoon$^{*}$ \\
Data Science and Artificial Intelligence Laboratory \\
ECE, Interdisciplinary Program in AI, and Institute of Engineering Research \\
Seoul National University, Seoul 08826, South Korea \\
\texttt{\{tuslkkk, t080205, kce407, sryoon\}@snu.ac.kr}\\
}

\date{}

\begin{document}
\maketitle

\begin{abstract}
To demystify the ``black box" property of deep neural networks for natural language processing (NLP), several methods have been proposed to interpret their predictions by measuring the change in prediction probability after erasing each token of an input.
Since existing methods replace each token with a predefined value (i.e., zero), the resulting sentence lies out of the training data distribution, yielding misleading interpretations.
In this study, we raise the out-of-distribution problem induced by the existing interpretation methods and present a remedy; we propose to marginalize each token out. 
We interpret various NLP models trained for sentiment analysis and natural language inference using the proposed method.


\end{abstract}

\section{Introduction}\label{sec::introduction}
The advent of deep learning has greatly improved the performances of natural language processing (NLP) models.
Consequently, the models are becoming more complex \cite{yang2019xlnet, liu2019roberta}, rendering it difficult to understand the \red{rationale} behind their predictions.
To use deep neural networks (DNNs) for making high-stakes decisions, the interpretability must be guaranteed to instill the trust in the public.
Hence, various attempts have been undertaken to provide an interpretation along with a prediction~\citep{gilpin2018explaining}.
\let\thefootnote\relax\footnotetext{*Correspondence to: Sungroh Yoon (sryoon@snu.ac.kr)}

\begin{figure}[t]
    \centering
    \includegraphics[width=\columnwidth]{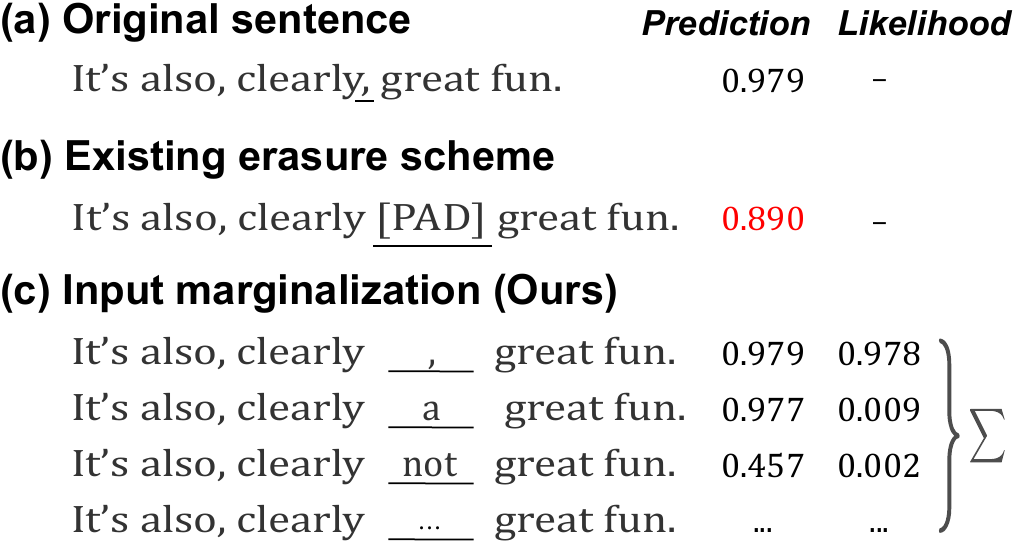}
    \caption{Given the original sentence (a), the existing erasure scheme (b) replaces each token with zero, i.e., [PAD] token. Our method (c) marginalizes each token out considering the likelihoods of candidate tokens.}
    \label{fig:ood_example}
    \vspace{-5pt}
\end{figure}

Research in computer vision aims to interpret a target model by measuring attribution scores, i.e., how much each pixel in an input image contributes to the final prediction \cite{simonyan2013deep, arras2017explaining, zeiler2014visualizing, lundberg2017unified}.
Since a pixel of an image corresponds to a token in a sentence, the attribution score of each token can provide an insight into the NLP model's internal reasoning process.
A straightforward approach is to ask, ``How would the model reaction change if each token was not there?'' and the change can be measured by the difference in softmax probabilities after \textit{erasing} each token. 
%
\citet{li2016understanding} proposed to erase each token by replacing it with a predefined value, i.e., zero.
This became a representative method for interpreting NLP models, followed by several papers using the similar erasure scheme~\citep{feng2018pathologies, prabhakaran2019perturbation, jin2019towards}. 

However, such an erasure scheme can cause out-of-distribution (OOD) problem, where the erased sentence deviates from the target model's training data distribution.
DNNs tend to assign a lower prediction probability to OOD samples than in-distribution samples \citep{hendrycks2016baseline}, as shown in Fig.~\ref{fig:ood_example}, which results in overestimated contribution of an unimportant token.
%
%
The OOD problem induced by the existing erasure scheme makes it difficult to identify whether high-scoring tokens actually contribute significantly to the prediction. 
%
%
In computer vision, several studies have highlighted the problem and attempted to address it~\citep{zintgraf2017visualizing, chang2018explaining, yi2020informationtheoretic}. 
To the best of our knowledge, the OOD problem has not been raised in the field of NLP, hence no solution has been suggested yet. 

In this study, we ask instead; ``How would the model react differently if there were other tokens instead of each token?", as proposed by~\citet{chang2018explaining, yi2020informationtheoretic}. 
We propose to marginalize each token out to mitigate the OOD problem of the existing erasure scheme. 
During the marginalization, our method measures the contribution of all probable candidate tokens considering their likelihoods. 
To calculate the likelihoods, we use the masked language modeling (MLM) of bidirectional encoder representations from transformers (BERT)~\cite{devlin2019bert}. 

Our contributions are as follows:
\begin{enumerate}[leftmargin=*]
\setlength\itemsep{0pt}
\item{To the best of our knowledge, we first raise the OOD problem that can arise when interpreting NLP models through the existing erasure schemes.}
\item{To avoid the OOD problem, we propose a new interpretation method, i.e., input marginalization using MLM for likelihood modeling.}
\item{We apply the proposed method to interpret various NLP models and quantitatively verify the correctness of the resulting interpretation.}
\end{enumerate}

\section{Related Works}\label{sec::related_works}
\subsection{Interpretation of NLP models}
\vspace{2pt}
Model-aware interpretation methods for DNNs use model information such as gradients. 
Saliency map \cite{simonyan2013deep} interprets an image classifier by computing the gradient of a target class logit score with respect to each input pixel. 
Since a token index is not ordinal as an image pixel, the gradient with respect to a token is meaningless.
Hence, \citet{li2016understanding} computed the gradient in an embedding space and \citet{arras2017explaining} distributed the class score to input embedding dimensions through layer-wise relevance propagation.
Both methods sum up the scores of each embedding dimension to provide the attribution score of a token.
Because the score can have a negative or positive sign, the sum may offset each other, so the contribution of the token may become zero even if it does contribute to the prediction.

Recently, the attention mechanism has been widely adopted to various NLP tasks \cite{bahdanau2014neural,vaswani2017attention,zhang2018deep} and there have been attempts to use the attention score as an interpretation. \cite{jain2019attention}.
However, its validity is still controversial~\cite{wiegreffe2019attention}. 

\vspace{2.5pt}
Model-agnostic approaches aim to interpret any types of model with no information other than its feed-forward outputs. 
They observe how much the prediction changes after erasing each unit of input. 
If it differs significantly, then the unit obtains a high attribution score. 
In computer vision, the measurement of prediction difference varies from the subtraction of probabilities~\cite{zeiler2014visualizing} to a log-odds probability difference~\cite{zintgraf2017visualizing}.
In the field of NLP, \citet{li2016understanding} interpreted NLP models by erasing each dimension of the embedding vector or the token itself, where the erasure was implemented by simply setting the value to a predefined value, i.e., zero. 
Such an erasure scheme can push the embedding vector or the input out of the training data distribution, thereby resulting in an inaccurate interpretation.

\subsection{Interpretation without OOD problem}
Several interpretation methods to mitigate the OOD problem have been proposed in computer vision.
\citet{zintgraf2017visualizing} proposed to marginalize each pixel out by assuming that the pixel value follows a Gaussian distribution.
It had limitations in that the Gaussian distribution differed from the real pixel distribution.
\citet{chang2018explaining} improved it by replacing an image segment with a plausible values generated from a deep generative model. 
\red{\citet{yi2020informationtheoretic} proposed to adopt an additional DNN to model the pixel distribution, which motivated our work the most.}

The method recently proposed by \citet{jin2019towards} may appear similar to ours as it marginalizes context words out to obtain the context-free attribution of a token. 
However, it still cannot overcome the OOD problem because it replaces the token with zero, similar to the existing methods. 
To the best of our knowledge, no attempt has been undertaken to raise and overcome the OOD problem that arises when interpreting NLP models.

\subsection{MLM of BERT}
BERT \cite{devlin2019bert}, one of the state-of-the-art natural language representations, is trained with two pre-training tasks: MLM and next sentence prediction. 
The MLM aims to infer the probability of a token to appear in the masked position of an input. 
As BERT is deeply bidirectional, it can consider the entire context of the sentence which enables the exact likelihood modeling.  
The likelihoods of the candidate tokens for marginalization are easily and accurately attainable using the MLM of BERT.

\section{Methods}\label{sec::methods}
We propose input marginalization to mitigate the OOD issue.
In the following subsections, we measure the attribution score using the weight of evidence and marginalize over all possible candidate tokens using the MLM of BERT.
We extend the method to multi-token cases and introduce adaptively truncated marginalization for an efficient computation. 
Finally, we propose a new metric, $\text{AUC}_\text{rep}$, to evaluate the proposed method faithfully.
The overall algorithm is provided in Algorithm \ref{pseudocode:truncated marginalization}.

\subsection{Measurement of model output difference} \label{subsection:WoE}
\red{To measure the changes in the model output, we adopt the widely used weight of evidence (WoE)} \cite{robnik2008explaining}, which is a log odds difference of prediction probabilities.
We define $\theta$ as the target model parameter, $y_c$ as a target class to be explained, and $\boldsymbol{x}$ as an input sentence.
We introduce $\boldsymbol{x}_{-i}$, i.e., $\boldsymbol{x}$ without $i$-th token $x_i$, to quantify the contribution of $x_i$ to predicting $y_c$.
WoE is formulated as follows:
\begin{equation} \label{eq:woe}
    \begin{aligned}
    \textrm{WoE}_{\theta, i}(y_c|\boldsymbol{x}) & = \textrm{log}_2(\textrm{odds}_{\theta}(y_c|\boldsymbol{x})) \\
    & -\textrm{log}_2(\textrm{odds}_{\theta}(y_c|\boldsymbol{x}_{-i})),
    \end{aligned}
\end{equation}
where $\textrm{odds}_{\theta}(y_c|\boldsymbol{x}) = p_{\theta}(y_c|\boldsymbol{x}) / (1-p_{\theta}(y_c|\boldsymbol{x}))$. 
$p_{\theta}(y_c|\boldsymbol{x}_{-i})$ captures the notion of the model response without the $i$-th token.
The first term of Eq.~\ref{eq:woe} can be easily obtained as it is the original prediction probability, while the second term is computed by input marginalization. 

\subsection{Input marginalization} \label{subsection:marginalization}
We rewrite the term $p(y_c| \boldsymbol{x}_{-i})$ of Eq. \ref{eq:woe} using marginalization as follows:
\begin{equation}\label{eq:marginalization}
    \begin{aligned}
        p(y_c|\boldsymbol{x}_{-i}) &= \sum_{\xti\in{\mathcal{V}}}p(y_c, \xti|\boldsymbol{x}_{-i}) \\
        &=\sum_{\xti\in{\mathcal{V}}}p(y_c|\xti,\boldsymbol{x}_{-i}) \cdot p(\xti|\boldsymbol{x}_{-i}).
    \end{aligned}
\end{equation}
Here, $\xti$ is a candidate token that can appear instead of $x_i$, and $\mathcal{V}$ is a set of vocabulary.
$p(y_c|\xti,\boldsymbol{x}_{-i})$ can be easily obtained by a single feed forward to the target model with the $i$-th token replaced with $\xti$.
We compute $p(\xti|\boldsymbol{x}_{-i})$, the likelihood of $\xti$ appearing in the $i$-th position, by substituting the $x_i$ with the ``[MASK]" token and feed forwarding it to BERT. 
The process of computing the attribution score of a token is repeated for all tokens in the sentence.
\begin{algorithm}[t]
    \caption{Input marginalization}
    \label{pseudocode:truncated marginalization}
    \begin{algorithmic}
        \State $\textbf{Input} \text{ Target model } \theta, \text{input } \boldsymbol{x}, \text{vocabulary } \mathcal{V}, $
        \State $\text{likelihood threshold } \sigma,  \text{ and target class } y_c$

        \State$\textbf{Output } \text{Attribution score } \boldsymbol{a}$
        \For {$\text{ }i=0 $ \textbf{ to } length($\boldsymbol{x}$)}
            \State $ m \leftarrow 0$ \Comment{Initialize attribution score}
            \State $ \boldsymbol{s} \leftarrow \text{copy } \boldsymbol{x} $
            \State $ \boldsymbol{s}_{i} \leftarrow \text{``[MASK]" token} $
            \ForAll {$\tilde{s}_i$ \text{in} $\mathcal{V}$}
            \State $ p(\tilde{s}_i|\boldsymbol{s}_{-i}) \leftarrow \text{BERT}_\text{MLM}(\boldsymbol{s}) $

            \If {$p(\tilde{s}_i|\boldsymbol{s}_{-i}) > \sigma$ }

                \State $ \boldsymbol{s}_i \leftarrow \tilde{s}_i $
                \State $m \leftarrow m + p(\tilde{s}_i|\boldsymbol{s}_{-i}) \cdot p_\theta(y_c|\boldsymbol{s})$ 
            \EndIf
            \EndFor
            \State $ \boldsymbol{a}_i = \text{logodds}_{\theta}(y_c|\boldsymbol{x})-\text{logodds}_{\theta}(m)  $ \\ \Comment{Prediction difference measurement}
        \EndFor
    \end{algorithmic}
\end{algorithm}
\subsection{Multi-token marginalization} \label{subsection:multi-token}
We can compute the attribution score for multiple tokens similarly. 
Let us assume that we wish to measure the joint contribution of two tokens $x_i$ and $x_j$. 
Eq. \ref{eq:marginalization} then becomes 
\begin{equation}\label{eq:multi-token}
        p(y_c|\boldsymbol{x}_{-i,j}) = \sum_{\xtj\in{\mathcal{V}}}\sum_{\xti\in{\mathcal{V}}}p(y_c, \xti, \xtj|\boldsymbol{x}_{-i,j}).
\end{equation}
Applying Bayes' theorem, $p(y_c, \xti, \xtj|\boldsymbol{x}_{-i,j})$ becomes $p(y_c|\xti,\xtj,\boldsymbol{x}_{-i,j}) \cdot p(\xti, \xtj|\boldsymbol{x}_{-i,j})$. 
The latter term of the multiplication can be decomposed into the multiplication of $p(\xti|\boldsymbol{x}_{-i}, \xtj)$ and $p(\xtj|\boldsymbol{x}_{-i,j})$.
Each term can be easily obtained by masking the corresponding position and feed-forwarding it to BERT even when $x_i$ and $x_j$ are distant. 
For more than two tokens, the attribution scores can be obtained in the similar way. \begin{figure*}[ht!]
    \centering
    \includegraphics[width=\textwidth]{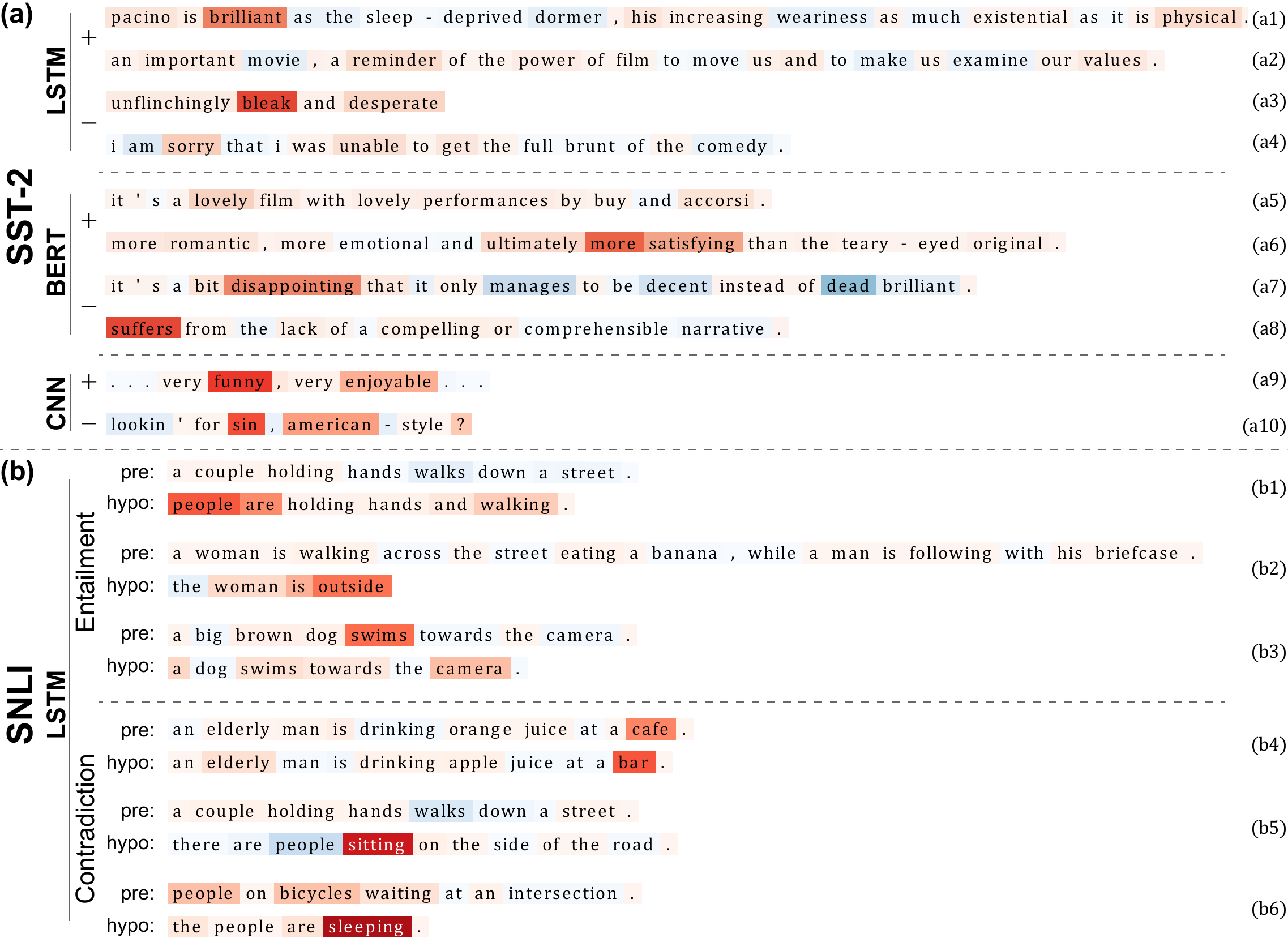}
    \caption{Interpretation results of the proposed method. ``+" and ``-" in (a) denote the positive and negative classes of the depicted sentences. ``pre" and ``hypo" in (b) denote premise and hypothesis of SNLI, respectively. Red and blue colors denote positive and negative contributions to the denoted classes, respectively.}
    \label{fig:interpretation_large}
\end{figure*}

\subsection{Adaptively truncated marginalization} \label{subsection:truncation}
%
The computational complexity for obtaining an attribution score of one token is $O(\abs{\mathcal{V}})$, where $\abs{\mathcal{V}}$ is the size of a vocabulary set.
For the tokenizer used in BERT, $\abs{\mathcal{V}}$ is greater than 30,000, and the same number of marginalization is required, which is computationally burdensome. 
For the efficient computation, we propose adaptively truncated marginalization.
If the magnitude of $p(\xti|\boldsymbol{x}_{-i})$ is insignificantly small, the contribution of $p(y_c|\xti, \boldsymbol{x}_{-i})$ to the summation in Eq.~\ref{eq:marginalization} becomes negligible.
Therefore, we marginalize only over candidates whose likelihoods are greater than a likelihood threshold ($\sigma$) and normalize the score.
Adaptively truncated marginalization approximates Eq.~\ref{eq:marginalization} as follows:
\vspace{-5pt}
\begin{equation}\label{eq:truncated_marg}
    \begin{aligned}
        p(y_c|\boldsymbol{x}_{-i}) \approx \frac{\sum_{\xti\in{\red{\tilde{\mathcal{V}}}}}p(y_c|\xti,\boldsymbol{x}_{-i}) \cdot p(\xti|\boldsymbol{x}_{-i})}{ \sum_{\xti\in\tilde{\mathcal{V}}}p(\xti|\boldsymbol{x}_{-i})},
    \end{aligned}
\end{equation}
\red{where $\tilde{\mathcal{V}}=\Set{\tilde{x_i}\in\mathcal{V}}{p(\tilde{x_i}|\boldsymbol{x}_{-i})>\sigma}$}.

Since the likelihood distributions depend on a token's position in the sentence, the number of marginalization varies for every $i$.
We will demonstrate the efficiency of adaptively truncated marginalization and find an optimal $\sigma$ in Section~\ref{sec::results}.

\subsection{Evaluation of interpretation} \label{subsection:evaluation_metric}
Inspired by \citet{petsiuk2018rise} and \citet{chang2018explaining}, we propose a metric $\textrm{AUC}_{\textrm{rep}}$ to evaluate interpretation methods for NLP models. 
Given the attribution scores of a sentence, \citet{petsiuk2018rise} plotted a prediction probability curve as pixels filled with zero in the order of importance.
If the interpretation is faithful, then the curve will drop rapidly, resulting in a small area under a curve (AUC). 
However, replacing the token with zero or removing it from a sentence can cause the OOD problem again.
Instead, we replace it with a token sampled from the distribution inferred by BERT MLM, as \citet{chang2018explaining} gradually replaced image segments with a generated sample. %
As MLM is trained by masking only a part of the input sentence, replacing too many tokens can degrade its modeling performance. 
Therefore, we calculate the $\textrm{AUC}$ until 20\% of the tokens are replaced, and refer to it as $\textrm{AUC}_{\textrm{rep}}$.

\section{Experimental Results}\label{sec::results}
\subsection{Experimental setup}
To show the model-agnostic and task-agnostic property of our method, we present interpretations of several types of DNNs trained for two tasks: sentiment analysis and natural language inference.

\vspace{2.5pt}
\textbf{SST-2} For sentiment analysis, we used the Stanford Sentiment Treebank binary classification corpus (SST-2) \citep{socher2013recursive}, which is a set of movie reviews labeled as positive or negative. 
We trained an 1-dimensional convolutional neural networks (CNNs) and a bidirectional long short-term memory (LSTM) with attention mechanism, and fine-tuned BERT~\cite{devlin2019bert}.
\begin{figure*}[t]
    \centering
    \includegraphics[width=\textwidth]{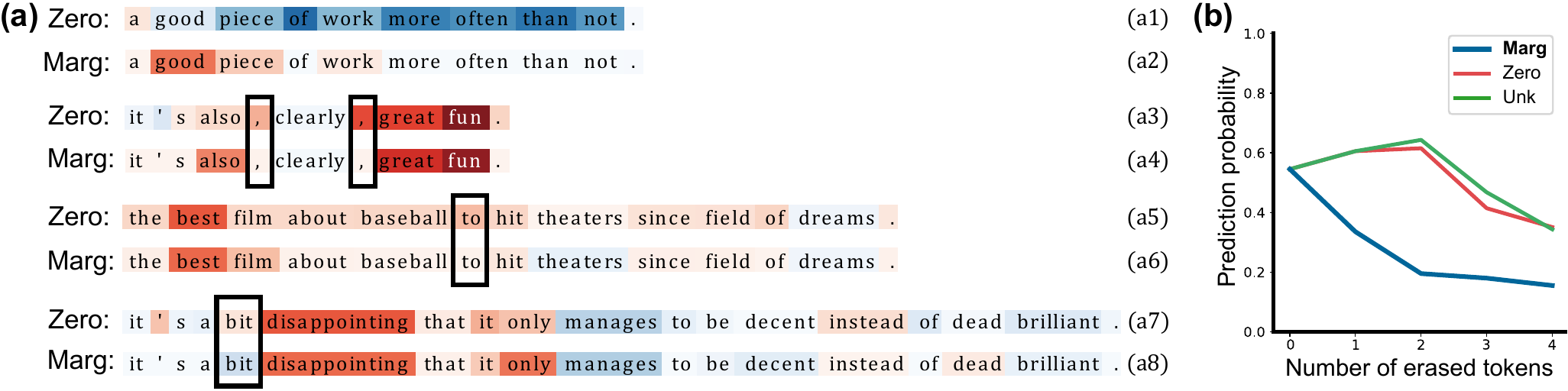}
    \caption{(a) shows examples of interpretations obtained by zero erasure and input marginalization (ours). Red and blue colors denote positive and negative contributions to the predicted classes, respectively. (a1-a6) are correctly classified as positive, and (a7-a8) are correctly classified to negative. (b) shows deletion curves of input marginalization, zero erasure, and ``[UNK]" erasure, which are abbreviated as ``Marg", ``Zero", and ``Unk", respectively. }
    \label{fig:zero_comparison}
\end{figure*}

\vspace{2.5pt}
\textbf{SNLI} For natural language inference, we used the Stanford natural language inference (SNLI) corpus \citep{young2014image}, a collection of pairs of two sentences, premise and hypothesis, annotated with three relationships between them: entailment, contradiction, and neutral. 
We trained the bidirectional LSTM for SNLI.

\vspace{2.5pt}
The final test accuracy of the target models is provided in Table \ref{table:modelaccuracy}. 
Note that the architectures of LSTM used for SST-2 and SNLI are distinct. 
Please refer to the Appendix for the detailed descriptions.
Throughout the experiments, we used the same tokenizer as BERT, where the zero-th token is ``[PAD]".
After training the target models, we interpreted their predictions through the proposed input marginalization.
\red{We used pre-trained BERT \cite{Wolf2019HuggingFacesTS} for likelihood modeling and $\sigma$ was set to $10^{-5}$}.

\begin{table}[h!]
\centering
\caption{Test accuracy of the target models}
\begin{tabular}{l|c|c|c}
\toprule
\multirow{2}{*}{\textbf{Corpus}}       & \multicolumn{3}{c}{\textbf{Model}} \\
                        & LSTM     & BERT     & CNN        \\
\toprule
SST-2           & 0.7753    & 0.8578 & 0.7300     \\ 
SNLI            & 0.6314    & \_      & \_          \\ 
\bottomrule
\end{tabular}
\label{table:modelaccuracy}
\end{table}

\subsection{Interpretation results} \label{subsection:interpretation_results}
The interpretation results of the proposed method are shown in Fig.~\ref{fig:interpretation_large}.
The color indicates the contribution of each token to the final prediction, with blue and red representing a negative and positive contribution, respectively.
Its intensity represents the magnitude of the attribution score.
\red{More examples are provided in Appendix.}

Fig.~\ref{fig:interpretation_large} (a) shows the interpretations of correct predictions for SST-2. 
The labels are shown in front of the sentences. 
For predicting the positive class, affirmative tokens such as ``brilliant" and ``funny" were attributed highly (a1, a9); if they are replaced with other tokens, the prediction probability will decrease significantly. 
Likewise, negative tokens such as ``disappointing" and ``suffers" were highlighted for predicting the negative class (a7, a8).
If positive and negative tokens appear in one sentence simultaneously, our method successfully assigned the opposite scores to those tokens: positive score to ``disappointing" and negative score to ``dead brilliant" for predicting negative class (a7).

The interpretations of the LSTM for SNLI are shown in Fig.~\ref{fig:interpretation_large} (b).
The sentences were correctly classified to the denoted class.
For predicting a class entailment, the token with a similar meaning were assigned high attribution score, such as ``swim" (b3). 
In contrast, tokens that makes two sentences contradicting were highlighted for predicting contradiction, such as ``cafe" vs. ``bar" (b4).


\subsection{Comparison to the existing erasing scheme}
In this section, we compare our method with the existing method proposed by \citet{li2015visualizing} through interpretations of models for SST-2. 
We refer to the existing method as zero erasure throughout the experiments as it replaces tokens with zero. 

\vspace{2.5pt}
\textbf{Qualitative comparison}
Interpretation results using input marginalization (Marg) and zero erasure (Zero) are depicted in Fig.~\ref{fig:zero_comparison} (a). 
Fig.~\ref{fig:zero_comparison} (a1-a6) and (a7-a8) were classified to positive and negative class, respectively. 
As shown in the figure, zero erasure often completely failed to interpret the prediction (a1). 
Zero erasure also assigned high attribution scores to uninformative tokens such as punctuation and ``to" (a3-a6). 
Our method showed clearer interpretations where unimportant tokens were given low attribution scores, while correctly highlighting the important ones. 
Moreover, the negative attribution was captured better than zero erasure. 
For example, in Fig.~\ref{fig:zero_comparison} (a7-a8), the token ``bit" reduces the degree of negativity of ``disappointing". 
Compared to potentially more assertive tokens (e.g. ``very"), the specific token diluted the negative sentiment of the sentence. 
The negative contribution of ``bit" to predicting the class ``negative" was captured only with our method.

\vspace{2.5pt}
\textbf{Quantitative comparison using $\text{AUC}_\text{rep}$}
We quantitatively compared our method with zero erasure using the $\text{AUC}_\text{rep}$ proposed in Section~\ref{sec::methods}.
Another baseline using ``[UNK]" token instead of zero was tested to verify that the OOD problem occurs no matter what predefined value is used. 
We would like to clarify that we did not consider the ``[MASK]" token because it is a special token dedicated for the pre-training of BERT.
It will obviously cause the OOD problem because it never appears during the training of target classifiers.

The deletion curves in Fig.~\ref{fig:zero_comparison} (b) shows the change in prediction probabilities as tokens with high attribution score are gradually replaced. 
The curves show that the deletion curve drawn using our method dropped more rapidly compared to the zero and ``[UNK]" erasures. 
The average $\text{AUC}_\text{rep}$ values for 700 SST-2 sentences are provided in Table~\ref{table:auc}, and the proposed method showed the lowest $\textrm{AUC}_{\textrm{rep}}$.
This result demonstrates that our method more accurately captures the importance of tokens than the existing erasure scheme. 
\begin{table}[h]
\centering
\caption{Comparison of $\text{AUC}_\text{rep}$ with the existing erasure scheme (the lower the better).}
\begin{tabular}{l|c|c|c}
\toprule
\multirow{2}{*}{ }       & \multicolumn{3}{c}{\textbf{Interpretation method}} \\
                        & Zero     & Unk     & Ours               \\
\toprule
$\textbf{AUC}_{\textbf{rep}}$           & 0.5284    & 0.5170     & \textbf{0.4972}      \\ 
\bottomrule
\end{tabular}
\label{table:auc}
\end{table}

\begin{figure*}[ht!] 
    \centering
    \includegraphics[width=\textwidth]{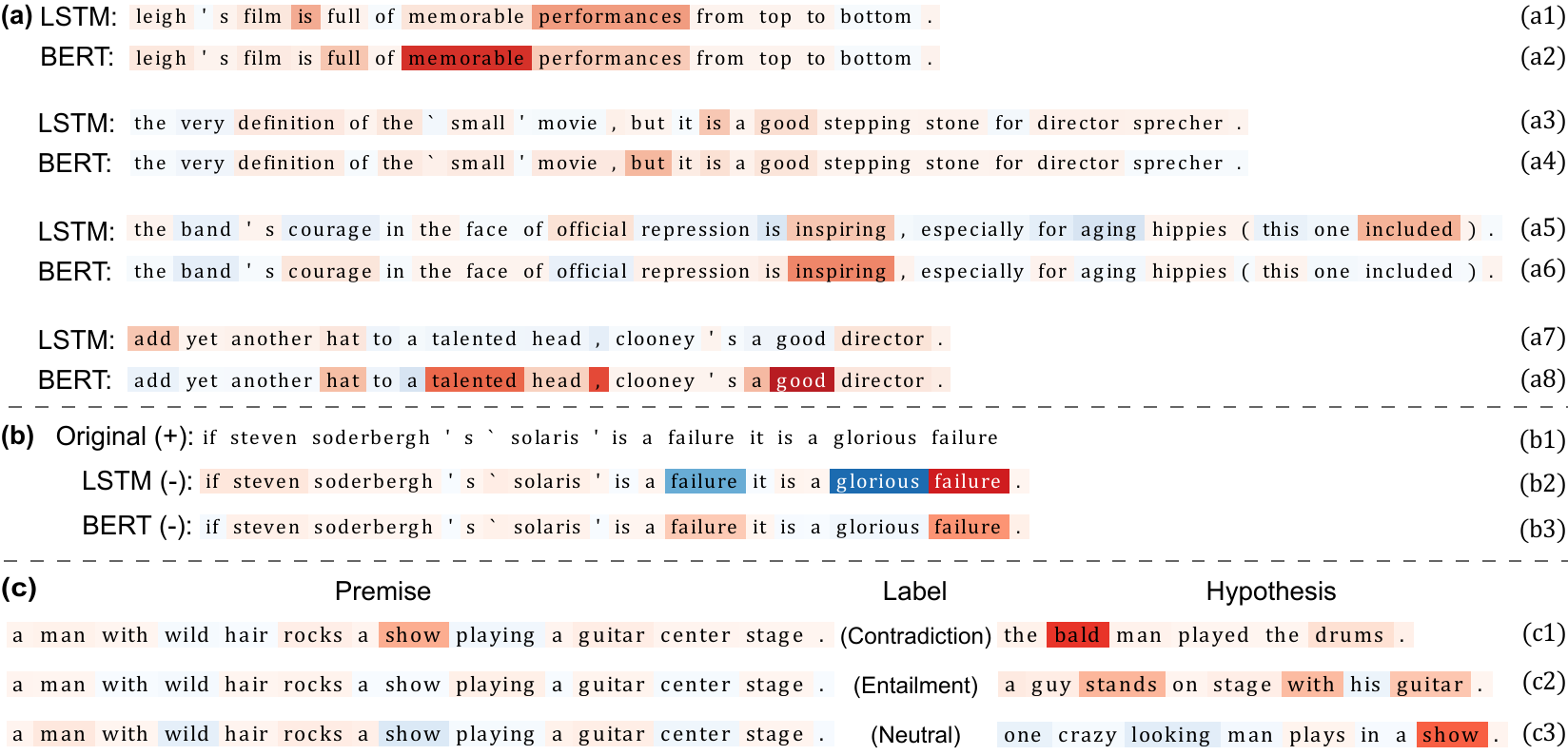}
    \caption{Interpretation results using input marginalization. Red and blue colors denote positive and negative contributions to the predicted classes. (a) shows interpretations of SST-2 predictions. (a1-a6) are correctly classified to positive, and (a7-a8) are correctly classified to negative. (b) shows positive sentences which are misclassified to negative by both LSTM and BERT. (c) shows the interpretations of SNLI predictions.}
    \label{fig:figure_additional}
\end{figure*}
 
\vspace{2.5pt}
\textbf{Quantitative comparison using SST-2 tags}
The SST-2 corpus provides not only sentence-level labels, but also five-class word-level sentiment tags ranging from very negative to very positive.
We can verify the validity of the attribution scores by comparing them with the word-level tags.
For simplicity, we merged very positive and positive, very negative and negative into positive (pos) and negative (neg), respectively, such that each token is given one tag among three.
If a sentence is correctly classified to positive, then three cases exist:
\begin{enumerate}[leftmargin=*]
\vspace{-5pt}
\setlength\itemsep{-3pt}
\item{pos-tagged word: contributes positively and significantly to the prediction}
\item{neut-tagged word: does not contribute much to the prediction}
\item{neg-tagged word: contributes negatively to the prediction},
\vspace{-5pt}
\end{enumerate}
where neut denotes neutral.

To assess if our method can assign high score to case \romannumeral 1 ), we measured the intersection of tokens (IoT) between pos-tagged tokens and highly attributed tokens in one sentence, motivated by intersection of union (IoU) which is a widely used interpretation evaluation metric in the vision field~\citep{chang2018explaining}.
IoT is defined as $\abs{P \cap T}/\abs{P}$, where $P$ denotes a set of + tagged tokens, and $T$ denotes a set of top-10 highly attributed tokens.
The average IoT for 100 sentences was 0.72 and 0.64 for our method and zero erasure, respectively.
This demonstrates that the tokens assigned with the highest attribution scores by our methods are likely to have a significant impact on the sentiment annotation.

A faithful interpretation method is expected to assign a small attribution score for the tokens belonging to \romannumeral 2 ).
For 500 interpretations, the average attribution score of the neutral words was 0.053 and 0.175 with our method and zero erasure, respectively. 
With our method, the candidate tokens inducing the OOD problem like zero have an insignificant effect on the final attribution score because they are assigned relatively low likelihoods. 

\subsection{Additional analysis using input marginalization} \label{additional_anaylsis}
The experimental results above demonstrates that our method can provide faithful interpretations. 
It thus can be used to analyze DNNs.  
First, we can compare the rationale of various models by analyzing their interpretations.
Fig.~\ref{fig:figure_additional} (a1-a4) show the interpretations of SST-2 samples correctly classified to positive by both BERT and LSTM.
BERT tended to focus more on affirmative tokens such as ``full" and ``memorable" (a2) and successfully identified the role of the token ``but" (a4) where the sentiment is reversed after it from negative to positive. 
Fig.~\ref{fig:figure_additional} (a5-a8) show the interpretations of samples that are labeled as positive but misclassified as negative by LSTM. 
The decisions of LSTM were significantly influenced by the word ``included" and ``add" (a5, a7). 
In contrast, BERT correctly classified them as positive by focusing on  ``inspiring" and ``good". 


Our method enables debugging the model by analyzing the misclassification case.
Fig.~\ref{fig:figure_additional} (b) shows the sentences whose true labels are positive but incorrectly classified as negative by both models.
We measured the attribution score with respect to the negative class.
For both models, the word ``failure" was assigned significantly high attribution score indicating that both models failed to recognize the overall positive sentiment of the sentence by focusing on the negativity inherent in the word. 


Fig.~\ref{fig:figure_additional} (c) shows different attribution scores assigned to the same premise when the hypothesis changes. 
It is shown that the tokens in the hypotheses received higher scores than those in the premises.
In fact, they obtained attribution scores twice as high as those in the premises for 500 interpretations.
We can potentially conclude that the model was trained to pay more attention to hypothesis, since the SNLI corpus consists of repetitive premises and varying hypotheses.
Moreover, (c1) shows that even if there are two contradictory word pairs, ``wild hair"-``bald" and ``guitar"-``drum", the model focused more on the the former. 
Our method allows potential model debugging when the interpretation turns out to be counterintuitive.









\subsection{Effect of language modeling}
In Eq.~\ref{eq:marginalization}, an exact modeling of the likelihood $p(\xti|\boldsymbol{x}_{-i})$ is important for the accurate calculation of the attribution scores.  
Hence, the high agreement between the modeled and the real-world distributions will result in a more accurate interpretation. 
We analyzed the effect of the likelihood modeling capability on the accuracy of interpretation results. 
We tested three additional likelihood modeling: uniform distribution, prior probability, and fine-tuned BERT MLM.

\vspace{2.5pt}
\textbf{Uniform}
$p(\xti|\boldsymbol{x}_{-i}) = 1/\abs{\mathcal{V}}=1/30522$ in the case of BERT tokenizer.

\vspace{2.5pt}
\textbf{Prior}
$p(\xti|\boldsymbol{x}_{-i})=p(\xti)$, defined by counting the frequency of each token in the training data.

\vspace{2.5pt}
\textbf{Fine-tuned MLM}
We fine-tuned the MLM of BERT with the SST-2 dataset for two epochs ($\text{MLM}_\text{fine}$). 
\begin{figure}[t!]
    \centering
    \includegraphics[width=\columnwidth]{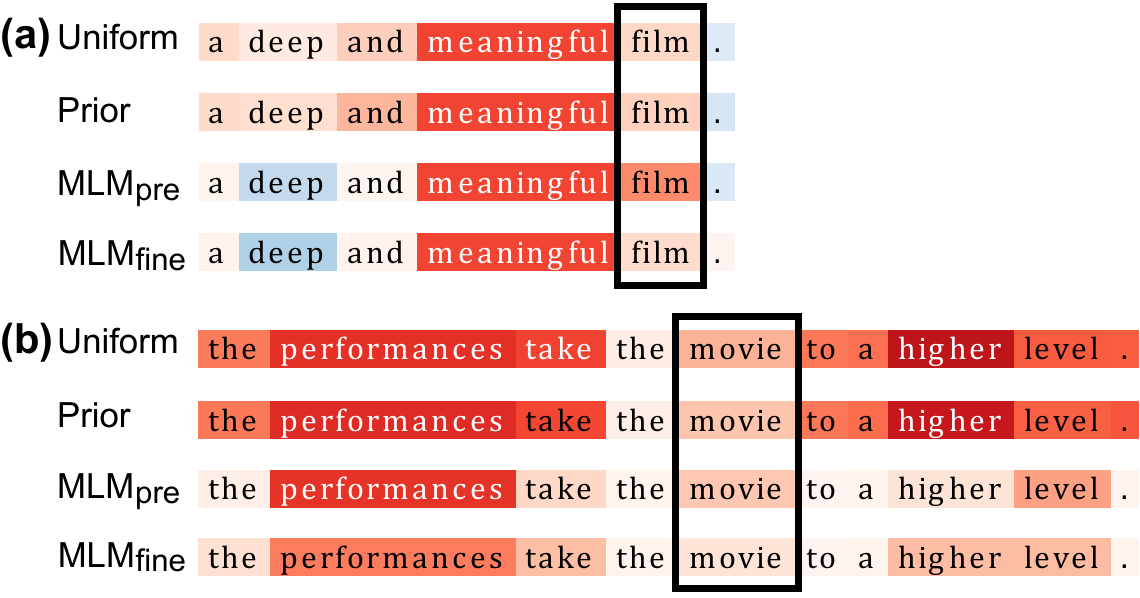}
    \caption{Interpretation results using different likelihood modelings. Each sentence is correctly classified to positive. Red and blue colors denote positive and negative contributions to the predicted classes.}
    \label{fig:lm_compare}
\end{figure}
\vspace{-2pt} \label{lm_compare}

\vspace{2.5pt}

Using each likelihood distribution modeling, we interpreted the BERT classifier trained for the SST-2 corpus.
The results are shown in Fig.~\ref{fig:lm_compare}.
The uniform distribution failed to provide an accurate interpretation. 
The result with prior probability modeling appeared slightly clearer, but was still misleading.
$\text{MLM}_\text{pre}$ successfully highlighted important tokens, but it assigned high scores to tokens that were not expected to contribute significantly to predicting the sentiment of a movie review (e.g., ``film" and ``movie" marked with box).
$\text{MLM}_\text{fine}$ yielded the most reasonable interpretation, where the attribution score of ``film" and ``movie" was reduced from 0.256 and 0.631 to 0.007 and 0.321, respectively, compared to $\text{MLM}_\text{pre}$.
We can expect the interpretation results to become more plausible as the likelihood modeling improves.

\subsection{Ablation study on adaptively truncated marginalization}
We introduced adaptively truncated marginalization in Section \ref{subsection:truncation} for a faster computation.
The full marginalization over all possible tokens yields the most exact attribution scores.
Thus, we searched for an optimal threshold $\sigma$ of adaptively truncated marginalization that reduces the computational complexity while maintaining a high correlation to the scores from full marginalization. 
We measured the correlation using the Pearson correlation coefficients.
Furthermore, we tested fixed truncation, which calculates top-$n$ likely candidates without considering the varying likelihoods depending on the position. 

Table~\ref{table:truncateablation} shows the Pearson correlation coefficient and the average number of marginalization under various thresholds. 
$\sigma=10^{-5}$ and $10^{-6}$ showed very similar interpretations to the full marginalization while the average marginalization number reduced to 2.5\% and 10.4\%, respectively, compared to 30,522 of the full marginalization.
We regarded $\sigma=10^{-5}$, which showed a lower number of marginalization under a similar correlation,  as the optimal value.
The fixed truncation showed a lower correlation under the similar average number of marginalization.
The computational complexity can be further reduced by accepting a slight loss in the accuracy.




\begin{table}[ht]
\centering
\caption{The Pearson correlation with full marginalization and the average number of marginalization under various thresholds. $\sigma$: likelihood threshold, $n$: marginalization number threshold for fixed truncation.} 
\begin{tabular}{l|l|c|c}
\toprule    
 \multicolumn{2}{c}{ }& \textbf{Corr}  & \textbf{Avg \#} \\
 \midrule
 \multirow{6}{*}{\rotatebox[origin=c]{90}{\textbf{Threshold}}}
 &$\sigma=10^{-6}$ & \textbf{0.9999}    & 3,186 \\
 &$\sigma=10^{-5}$ & \textbf{0.9999}    & 791 \\
 &$\sigma=10^{-4}$ & 0.9988    & 171 \\
 &$\sigma=10^{-3}$ & 0.9928    & 33 \\
 \cmidrule{2-4}
 &$n=10^3$  &0.9958    &1,000 \\
 &$n=10^2$      &0.9823  &100  \\

\bottomrule
\end{tabular}
\label{table:truncateablation}
\end{table}

\section{Conclusion}\label{sec::conclusion}
Interpretability is becoming more important owing to the increase in deep learning in NLP. 
Hence, several interpretation methods have been proposed, and we reviewed their limitations throughout the paper. 
Among them, we focused on the OOD problem arising from the widely used zero erasure scheme, which results in misleading interpretation. 
To the best of our knowledge, neither the OOD problem has been raised in interpreting NLP models nor the attempt to resolve it has been undertaken. 
Our proposed input marginalization, which can mitigate the OOD problem, can result in a faithful interpretation, thereby enabling better understanding of ``black box'' DNNs.

The scope of this study was primarily focused on interpreting DNNs for sentiment analysis and natural language inference. 
Regarding the model-agnostic and task-agnostic properties of our method, they are applicable to any types of NLP model for various tasks, such as neural machine translation and visual question answering.
It will be meaningful to interpret the state-of-the-art models such as XLNet~\cite{yang2019xlnet} and ELECTRA~\cite{clark2019electra}.
In addition, as experimentally analyzed, the interpretation result of our method is affected by the likelihood modeling performance.  
We can expect even more faithful interpretation if the modeling performance improves.

\section*{Acknowledgement}\label{sec::acknowledgement}
This work was supported by the National Research Foundation of Korea (NRF) grant funded by the Korea government (Ministry of Science and ICT) [2018R1A2B3001628], the Brain Korea 21 Plus Project in 2020, and Hyundai Motor Company.

\bibliographystyle{acl_natbib}
\bibliography{emnlp2020}

\begin{thebibliography}{28}
\expandafter\ifx\csname natexlab\endcsname\relax\def\natexlab#1{#1}\fi

\bibitem[{Arras et~al.(2017)Arras, Montavon, M{\"u}ller, and
  Samek}]{arras2017explaining}
Leila Arras, Gr{\'e}goire Montavon, Klaus-Robert M{\"u}ller, and Wojciech
  Samek. 2017.
\newblock Explaining recurrent neural network predictions in sentiment
  analysis.
\newblock \emph{EMNLP 2017}, page 159.

\bibitem[{Bahdanau et~al.(2014)Bahdanau, Cho, and Bengio}]{bahdanau2014neural}
Dzmitry Bahdanau, Kyunghyun Cho, and Yoshua Bengio. 2014.
\newblock Neural machine translation by jointly learning to align and
  translate.
\newblock \emph{arXiv preprint arXiv:1409.0473}.

\bibitem[{Chang et~al.(2018)Chang, Creager, Goldenberg, and
  Duvenaud}]{chang2018explaining}
Chun-Hao Chang, Elliot Creager, Anna Goldenberg, and David Duvenaud. 2018.
\newblock Explaining image classifiers by counterfactual generation.

\bibitem[{Clark et~al.(2019)Clark, Luong, Le, and Manning}]{clark2019electra}
Kevin Clark, Minh-Thang Luong, Quoc~V Le, and Christopher~D Manning. 2019.
\newblock Electra: Pre-training text encoders as discriminators rather than
  generators.
\newblock In \emph{International Conference on Learning Representations}.

\bibitem[{Devlin et~al.(2019)Devlin, Chang, Lee, and
  Toutanova}]{devlin2019bert}
Jacob Devlin, Ming-Wei Chang, Kenton Lee, and Kristina Toutanova. 2019.
\newblock Bert: Pre-training of deep bidirectional transformers for language
  understanding.
\newblock In \emph{Proceedings of the 2019 Conference of the North American
  Chapter of the Association for Computational Linguistics: Human Language
  Technologies, Volume 1 (Long and Short Papers)}, pages 4171--4186.

\bibitem[{Feng et~al.(2018)Feng, Wallace, Grissom~II, Iyyer, Rodriguez, and
  Boyd-Graber}]{feng2018pathologies}
Shi Feng, Eric Wallace, Alvin Grissom~II, Mohit Iyyer, Pedro Rodriguez, and
  Jordan Boyd-Graber. 2018.
\newblock Pathologies of neural models make interpretations difficult.
\newblock In \emph{Proceedings of the 2018 Conference on Empirical Methods in
  Natural Language Processing}, pages 3719--3728.

\bibitem[{Gilpin et~al.(2018)Gilpin, Bau, Yuan, Bajwa, Specter, and
  Kagal}]{gilpin2018explaining}
Leilani~H Gilpin, David Bau, Ben~Z Yuan, Ayesha Bajwa, Michael Specter, and
  Lalana Kagal. 2018.
\newblock Explaining explanations: An overview of interpretability of machine
  learning.
\newblock In \emph{2018 IEEE 5th International Conference on data science and
  advanced analytics (DSAA)}, pages 80--89. IEEE.

\bibitem[{Hendrycks and Gimpel(2016)}]{hendrycks2016baseline}
Dan Hendrycks and Kevin Gimpel. 2016.
\newblock A baseline for detecting misclassified and out-of-distribution
  examples in neural networks.
\newblock \emph{arXiv preprint arXiv:1610.02136}.

\bibitem[{Jain and Wallace(2019)}]{jain2019attention}
Sarthak Jain and Byron~C Wallace. 2019.
\newblock Attention is not explanation.
\newblock \emph{arXiv preprint arXiv:1902.10186}.

\bibitem[{Jin et~al.(2019)Jin, Du, Wei, Xue, and Ren}]{jin2019towards}
Xisen Jin, Junyi Du, Zhongyu Wei, Xiangyang Xue, and Xiang Ren. 2019.
\newblock Towards hierarchical importance attribution: Explaining compositional
  semantics for neural sequence models.
\newblock \emph{arXiv preprint arXiv:1911.06194}.

\bibitem[{Li et~al.(2015)Li, Chen, Hovy, and Jurafsky}]{li2015visualizing}
Jiwei Li, Xinlei Chen, Eduard Hovy, and Dan Jurafsky. 2015.
\newblock Visualizing and understanding neural models in nlp.
\newblock \emph{arXiv preprint arXiv:1506.01066}.

\bibitem[{Li et~al.(2016)Li, Monroe, and Jurafsky}]{li2016understanding}
Jiwei Li, Will Monroe, and Dan Jurafsky. 2016.
\newblock Understanding neural networks through representation erasure.
\newblock \emph{arXiv preprint arXiv:1612.08220}.

\bibitem[{Liu et~al.(2019)Liu, Ott, Goyal, Du, Joshi, Chen, Levy, Lewis,
  Zettlemoyer, and Stoyanov}]{liu2019roberta}
Yinhan Liu, Myle Ott, Naman Goyal, Jingfei Du, Mandar Joshi, Danqi Chen, Omer
  Levy, Mike Lewis, Luke Zettlemoyer, and Veselin Stoyanov. 2019.
\newblock Roberta: A robustly optimized bert pretraining approach.
\newblock \emph{arXiv preprint arXiv:1907.11692}.

\bibitem[{Lundberg and Lee(2017)}]{lundberg2017unified}
Scott~M Lundberg and Su-In Lee. 2017.
\newblock A unified approach to interpreting model predictions.
\newblock In \emph{Advances in neural information processing systems}, pages
  4765--4774.

\bibitem[{Petsiuk et~al.(2018)Petsiuk, Das, and Saenko}]{petsiuk2018rise}
Vitali Petsiuk, Abir Das, and Kate Saenko. 2018.
\newblock Rise: Randomized input sampling for explanation of black-box models.
\newblock \emph{arXiv preprint arXiv:1806.07421}.

\bibitem[{Prabhakaran et~al.(2019)Prabhakaran, Hutchinson, and
  Mitchell}]{prabhakaran2019perturbation}
Vinodkumar Prabhakaran, Ben Hutchinson, and Margaret Mitchell. 2019.
\newblock Perturbation sensitivity analysis to detect unintended model biases.
\newblock \emph{arXiv preprint arXiv:1910.04210}.

\bibitem[{Robnik-{\v{S}}ikonja and Kononenko(2008)}]{robnik2008explaining}
Marko Robnik-{\v{S}}ikonja and Igor Kononenko. 2008.
\newblock Explaining classifications for individual instances.
\newblock \emph{IEEE Transactions on Knowledge and Data Engineering},
  20(5):589--600.

\bibitem[{Simonyan et~al.(2013)Simonyan, Vedaldi, and
  Zisserman}]{simonyan2013deep}
Karen Simonyan, Andrea Vedaldi, and Andrew Zisserman. 2013.
\newblock Deep inside convolutional networks: Visualising image classification
  models and saliency maps.
\newblock \emph{arXiv preprint arXiv:1312.6034}.

\bibitem[{Socher et~al.(2013)Socher, Perelygin, Wu, Chuang, Manning, Ng, and
  Potts}]{socher2013recursive}
Richard Socher, Alex Perelygin, Jean Wu, Jason Chuang, Christopher~D Manning,
  Andrew~Y Ng, and Christopher Potts. 2013.
\newblock Recursive deep models for semantic compositionality over a sentiment
  treebank.
\newblock In \emph{Proceedings of the 2013 conference on empirical methods in
  natural language processing}, pages 1631--1642.

\bibitem[{Vaswani et~al.(2017)Vaswani, Shazeer, Parmar, Uszkoreit, Jones,
  Gomez, Kaiser, and Polosukhin}]{vaswani2017attention}
Ashish Vaswani, Noam Shazeer, Niki Parmar, Jakob Uszkoreit, Llion Jones,
  Aidan~N Gomez, {\L}ukasz Kaiser, and Illia Polosukhin. 2017.
\newblock Attention is all you need.
\newblock In \emph{Advances in neural information processing systems}, pages
  5998--6008.

\bibitem[{Wiegreffe and Pinter(2019)}]{wiegreffe2019attention}
Sarah Wiegreffe and Yuval Pinter. 2019.
\newblock Attention is not not explanation.
\newblock \emph{arXiv preprint arXiv:1908.04626}.

\bibitem[{Wolf et~al.(2019)Wolf, Debut, Sanh, Chaumond, Delangue, Moi, Cistac,
  Rault, Louf, Funtowicz, and Brew}]{Wolf2019HuggingFacesTS}
Thomas Wolf, Lysandre Debut, Victor Sanh, Julien Chaumond, Clement Delangue,
  Anthony Moi, Pierric Cistac, Tim Rault, R'emi Louf, Morgan Funtowicz, and
  Jamie Brew. 2019.
\newblock Huggingface's transformers: State-of-the-art natural language
  processing.
\newblock \emph{ArXiv}, abs/1910.03771.

\bibitem[{Yang et~al.(2019)Yang, Dai, Yang, Carbonell, Salakhutdinov, and
  Le}]{yang2019xlnet}
Zhilin Yang, Zihang Dai, Yiming Yang, Jaime Carbonell, Ruslan Salakhutdinov,
  and Quoc~V Le. 2019.
\newblock Xlnet: Generalized autoregressive pretraining for language
  understanding.
\newblock \emph{arXiv preprint arXiv:1906.08237}.

\bibitem[{Yi et~al.(2020)Yi, Kim, Kim, and Yoon}]{yi2020informationtheoretic}
Jihun Yi, Eunji Kim, Siwon Kim, and Sungroh Yoon. 2020.
\newblock \href {http://arxiv.org/abs/2009.11150} {Information-theoretic visual
  explanation for black-box classifiers}.

\bibitem[{Young et~al.(2014)Young, Lai, Hodosh, and
  Hockenmaier}]{young2014image}
Peter Young, Alice Lai, Micah Hodosh, and Julia Hockenmaier. 2014.
\newblock From image descriptions to visual denotations: New similarity metrics
  for semantic inference over event descriptions.
\newblock \emph{Transactions of the Association for Computational Linguistics},
  2:67--78.

\bibitem[{Zeiler and Fergus(2014)}]{zeiler2014visualizing}
Matthew~D Zeiler and Rob Fergus. 2014.
\newblock Visualizing and understanding convolutional networks.
\newblock In \emph{European conference on computer vision}, pages 818--833.
  Springer.

\bibitem[{Zhang et~al.(2018)Zhang, Wang, and Liu}]{zhang2018deep}
Lei Zhang, Shuai Wang, and Bing Liu. 2018.
\newblock Deep learning for sentiment analysis: A survey.
\newblock \emph{Wiley Interdisciplinary Reviews: Data Mining and Knowledge
  Discovery}, 8(4):e1253.

\bibitem[{Zintgraf et~al.(2017)Zintgraf, Cohen, Adel, and
  Welling}]{zintgraf2017visualizing}
Luisa~M Zintgraf, Taco~S Cohen, Tameem Adel, and Max Welling. 2017.
\newblock Visualizing deep neural network decisions: Prediction difference
  analysis.
\newblock \emph{arXiv preprint arXiv:1702.04595}.

\end{thebibliography}

\end{document}